\documentclass[10pt,twocolumn,letterpaper]{article}

\usepackage[pagenumbers]{cvpr} % To force page numbers, e.g. for an arXiv version
\usepackage[accsupp]{axessibility}

\usepackage{amsmath}
\DeclareMathOperator\arctanh{arctanh}

\usepackage{array}
\newcolumntype{L}{>{\raggedright\arraybackslash} p{0.10\textwidth}}
\usepackage[group-separator = {,}]{siunitx}

\definecolor{cvprblue}{rgb}{0.21,0.49,0.74}
\usepackage[pagebackref,breaklinks,colorlinks,allcolors=cvprblue]{hyperref}

\title{Fast Sphericity and Roundness approximation in 2D and 3D using Local Thickness}

\author{Pawel Tomasz Pieta\\
\and
Peter Winkel Rasmussen
\and
Anders Bjorholm Dahl
\and
Anders Nymark Christensen \\ \\
{\tt\small \{papi, pwra, abda, anym\}@dtu.dk}\\ 
Technical University of Denmark, 
Kgs. Lyngby, Denmark\\
}

\begin{document}
\maketitle
\begin{abstract}
Sphericity and roundness are fundamental measures used for assessing object uniformity in 2D and 3D images. However, using their strict definition makes computation costly. As both 2D and 3D microscopy imaging datasets grow larger, there is an increased demand for efficient algorithms that can quantify multiple objects in large volumes. We propose a novel approach for extracting sphericity and roundness based on the output of a local thickness algorithm. For sphericity, we simplify the surface area computation by modeling objects as spheroids/ellipses of varying lengths and widths of mean local thickness. For roundness, we avoid a complex corner curvature determination process by approximating it with local thickness values on the contour/surface of the object. The resulting methods provide an accurate representation of the exact measures while being significantly faster than their existing implementations.
\end{abstract}    
\section{Introduction}
\label{sec:intro}

The analysis of segmentation masks is a standard part of image processing pipelines. Especially within microscopy, the morphology of segmented objects is a frequent research target, requiring a wide range of descriptors~\cite{yin2014a,boquet-pujadas2017a,deng2020a,boquet-pujadas2021a,heertje1997a,kim1990a,dulaimi2018a,lesty1989a,kalinin2018a}. The simplest statistics and shape measures can be seamlessly extracted using standard image-processing libraries~\cite{opencv_library,van2014scikit,2020SciPy-NMeth}. While these are useful and fast to extract, they are often not descriptive enough for a detailed analysis of the shape of the object.

One of the more advanced shape measures is local thickness~\cite{hildebrand1997a_localthickness,imagej_localthickness}, defined as the radius of the largest circle/sphere that fits in the object at any point inside it. Although its calculation is relatively complex, recent advancements have shown that it can be efficiently accelerated while maintaining high accuracy~\cite{dahl2023a_localthickness}.

\begin{figure}[!t]
\centering
\includegraphics[width=0.95\columnwidth]{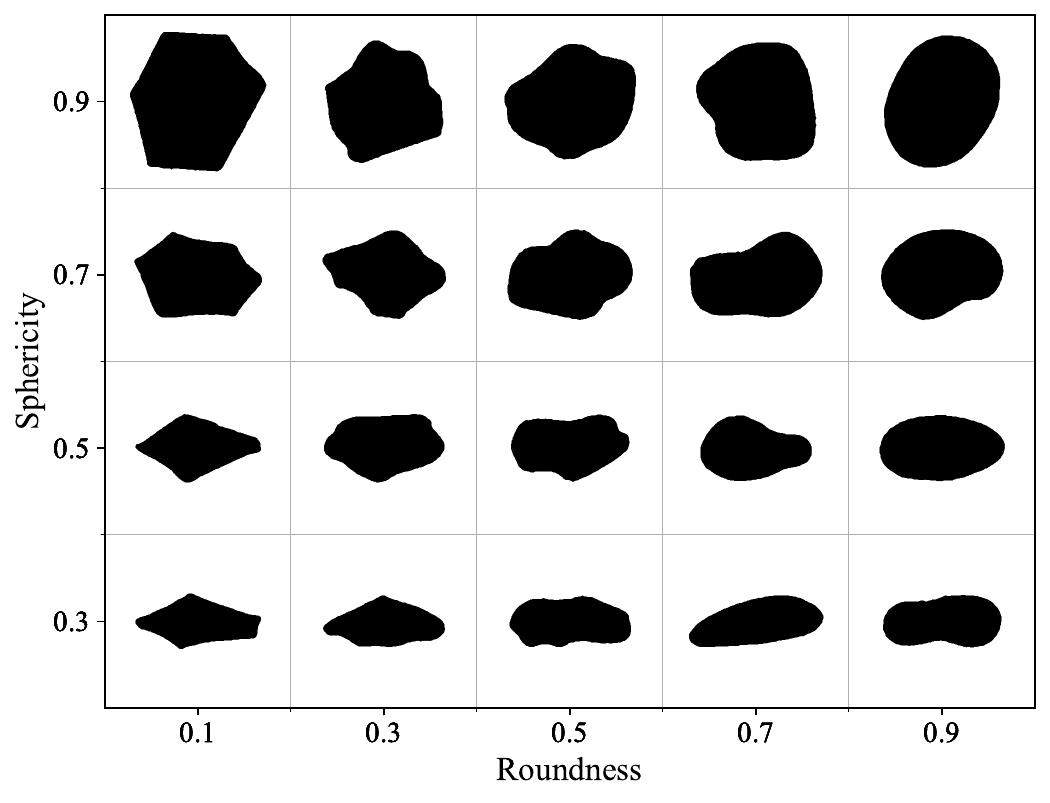}
\caption{Visual chart for estimating sphericity and roundness from 2D silhouettes, adapted from~\cite{krumbein1951a}.}
\label{fig:kumbrein_plot}
\end{figure}

Similar to local thickness --- sphericity and roundness can also serve as object descriptors, focusing much more on its shape than its dimensions. Originating from geology~\cite{wadell1932a}, they are widely used in the study of rock/granular particle shape~\cite{alrkaby2017a,beakawi2018a,goudarzy2018a,resentini2018a,suh2017a,vangla2018a}. However, their descriptive power extends their applicability beyond geology, finding use both within bio-imaging \cite{aguilar2021morphological,grexa2021spheroidpicker} and other image-based studies~\cite{han2018a,li2017a,papadikis2010a,vukadinovic2023a,hopman2023a}.

Both sphericity and roundness capture how close an object is to a perfect circle/sphere, but they describe two different, complementary characteristics. Sphericity depends primarily on the elongation of the object, while roundness captures the roughness or inconsistency of its edges. As a result, an elongated but smooth object will have low sphericity and high roundness, while a compact and rugged object will have high sphericity but low roundness (\cref{fig:kumbrein_plot}).

Many implementations of the two measures have been suggested, both for 2D and 3D data~\cite{fonseca2012a,windreich2003a,lin2005a,zheng2015a,zheng2021a,bullard2013a,zhao2016a,mat2019a}. Out of the two, sphericity is easier to implement, typically requiring an extraction of an object's perimeter in 2D and mesh in 3D. Calculating roundness continues to be a challenge, especially in 3D, as in the most strict form it requires collecting radii of curvature of all object corners and ridges. While some approximate methods for calculating roundness also exist~\cite{drevin2002a,roussillon2009a}, their implementations are rarely available. Overall, even the most popular implementations of the two measures are relatively complex and unfit for modern data-heavy analysis. They tend to perform adequately when evaluating a handful of objects, but often struggle to accommodate large groups of objects at once.

We propose a fast and scalable method for extracting the sphericity and roundness of objects from 2D and 3D masks based on their local thickness values. Noticing the parallels between the sphere-based definition of local thickness and inputs needed for calculating the two measures, we simplify their calculation to a set of fast operations that can easily be applied to all image objects at once. For sphericity, we model the objects as ellipses (in 2D) and spheroids (in 3D) with a width of mean local thickness and volume-dependent height. This approach lets us avoid the meshing step with a minimal loss in accuracy for the majority of the most common object shapes. For roundness, we extract local thickness values from a simplified perimeter/shell of the object and treat them as the curvature radii. While not exact, the resulting measure exhibits high levels of correlation to the original measure. Our implementation is directly available for use in Python, as a pip-installable package under \url{https://github.com/PaPieta/fast_rs}. 

\section{Background}
\label{sec:background}
The most established definitions of sphericity and roundness were formulated by \citet{wadell1932a, wadell1933a, wadell1935a}. Both measures were designed to describe the geometry of 3D objects, such as rocks, pebbles, or particles. While sphericity was a 3D definition from the start, roundness was only defined for 2D object projections. Initially, they were assessed primarily through physical measurements, but the complexity of these assessments led to the creation of charts that allowed for approximate visual estimation~\cite{krumbein1941a,krumbein1951a,powers1953a} (\cref{fig:kumbrein_plot}).

Later developments brought more sophisticated measurement techniques, such as using specialized tools~\cite{ASAHINA2011,hayakawa2005a,BHATTACHARYA2020} or performing image-based processing~\cite{yeralan1988computerized,diepenbroek1992round}. The simplicity and potential of the image-based approach, as well as advancements in 3D imaging, have fueled the research and development of new roundness and sphericity algorithms (detailed in ~\cref{sec:related_work}). These methods often used the sphericity and roundness charts as baselines for evaluating and benchmarking their performance.

\subsection{Sphericity background}

Sphericity is defined in 3D as \textit{the ratio of the surface area of a sphere, of the same volume as the particle, to the actual surface area of the particle} \cite{wadell1932a}:

\begin{equation}
    \mathcal{S}_{3D} = \frac{\pi^{\frac{1}{3}}(6 V)^{\frac{2}{3}}}{S}
\end{equation}
where $V$ is the volume of the particle/object and $S$ is its surface area.

A direct translation of 3D sphericity to 2D suggests the use of \emph{perimeter sphericity}~\cite{cox1927a,kuo2000a,altuhafi2013a}, or ISO(2008) circularity~\cite{ISO92766}:
\begin{equation}
\label{eq:s2d_p}
    \mathcal{S}_{2D,P} = \frac{P_c}{P_o}
\end{equation}
where $P_c$ is the perimeter of a circle having the same projected area as the object, and $P_o$ is the perimeter of the object.

However, it is not clear which of the many proposed specificity definitions~\cite{mitchell2005a,Rodriguez1008311} was used for preparing the baseline visual approximation charts. A recent investigation~\cite{zheng2015a} suggests that \citet{krumbein1951a} used an alternative definition of \emph{width-to-lenght ratio sphericity} in his charts (\cref{fig:kumbrein_plot}):
\begin{equation}
    \mathcal{S}_{2D,WL} = \frac{d_2}{d_1}
\end{equation}
where $d_1$ and $d_2$ are the length and width of the object.

\subsection{Roundness background}

Roundness is defined in 2D as \textit{the ratio of the average curvature radius of all corners of a particle to the radius of the maximum inscribed circle} \cite{wadell1932a}:
\begin{equation}
    \mathcal{R}_{2D} = \frac{\sum^{N}_{i=1} r_i}{NR}
\end{equation}
where $r_i$ is the curvature radius of the $i$-th corner, $R$ is the radius of the maximum inscribed circle and $N$ is the number of corners.

Importantly, roundness is often assessed alongside another property, surface roughness \cite{zheng2015a,shin2013a}. It describes how smooth the texture of the particle surface is, and because of that, it is highly scale-dependent. Depending on the chosen threshold, a protruding part of an object can either be classified as a corner for roundness calculation or a contributor to surface roughness.

Defining roundness in 3D is challenging because protrusions in the shape of a 3D object include not only corners but also ridges. As a result, translating roundness directly to 3D requires accounting for the mean curvature of both geometries. However, an exact approach for incorporating the ridge contribution varies heavily depending on the proposed method. A generic 3D roundness equation could be defined as:
\begin{equation}
    \mathcal{R}_{3D} = \frac{\overline{r}}{R}
\end{equation}
where $\overline{r}$ represents the mean curvature of corners and edges and $R$ is the radius of the maximum inscribed sphere.

\section{Related work}
\label{sec:related_work}

All 2D definitions of sphericity are relatively straightforward to implement and can be calculated -- either directly or with just a few commands -- using various programs and image processing libraries. For 3D sphericity, the primary challenge is measuring the object's surface area. Over the years, many approaches have been proposed for this task~\cite{fonseca2012a,windreich2003a}, but by far the most recognized one involves calculating the surface area of a meshed object, typically achieved with the marching cubes algorithm~\cite{lewiner2003a,lorensen1987a,lin2005a}. Although some recent contributions propose alternative methods for calculating 3D sphericity, these are often part of more complex modeling approaches, primarily aimed at addressing 3D roundness calculation~\cite{mat2019a,bullard2013a}.

Calculating the exact 2D roundness is challenging due to the added complexity of detecting shape corners. Some solutions have addressed this by relaxing the definition, quantifying the curvature of the entire edge rather than just the corners~\cite{drevin2002a,roussillon2009a}. While these measures no longer maintain the original \numrange{0}{1} range of roundness, they correlate strongly with it. \citet{zheng2015a} were the first to propose a direct calculation of 2D roundness that incorporated filtering out surface roughness, corner detection, and corner curvature calculation. Their algorithm successfully recreated Krumbein’s chart roundness measures with relatively small errors, making it the most recognized solution for 2D roundness calculation.

In 3D, the need to quantify the curvature of both ridges and corners further complicates roundness calculation. Building on the 2D method, \citet{zheng2021a} proposed an algorithm where 2D roundness is calculated on slices of the object's volume, only to be used for detecting/fitting spheres in the whole object. The ridge contribution is addressed by discretizing it into multiple corner-like elements. Other methods first mesh the object’s surface, identify ridge and corner regions, and integrate the curvature values from these areas~\cite{bullard2013a,zhao2016a}. Several alternative methods have also been proposed~\cite{mat2019a,li2023a}, but generally, all are yet to become the standard in the field, and none have an implementation available online. 
\section{Method}
\label{sec:method}

\begin{figure*}[!t]
\centering
     \begin{subfigure}[b]{0.228\textwidth}
         \includegraphics[width=\columnwidth]{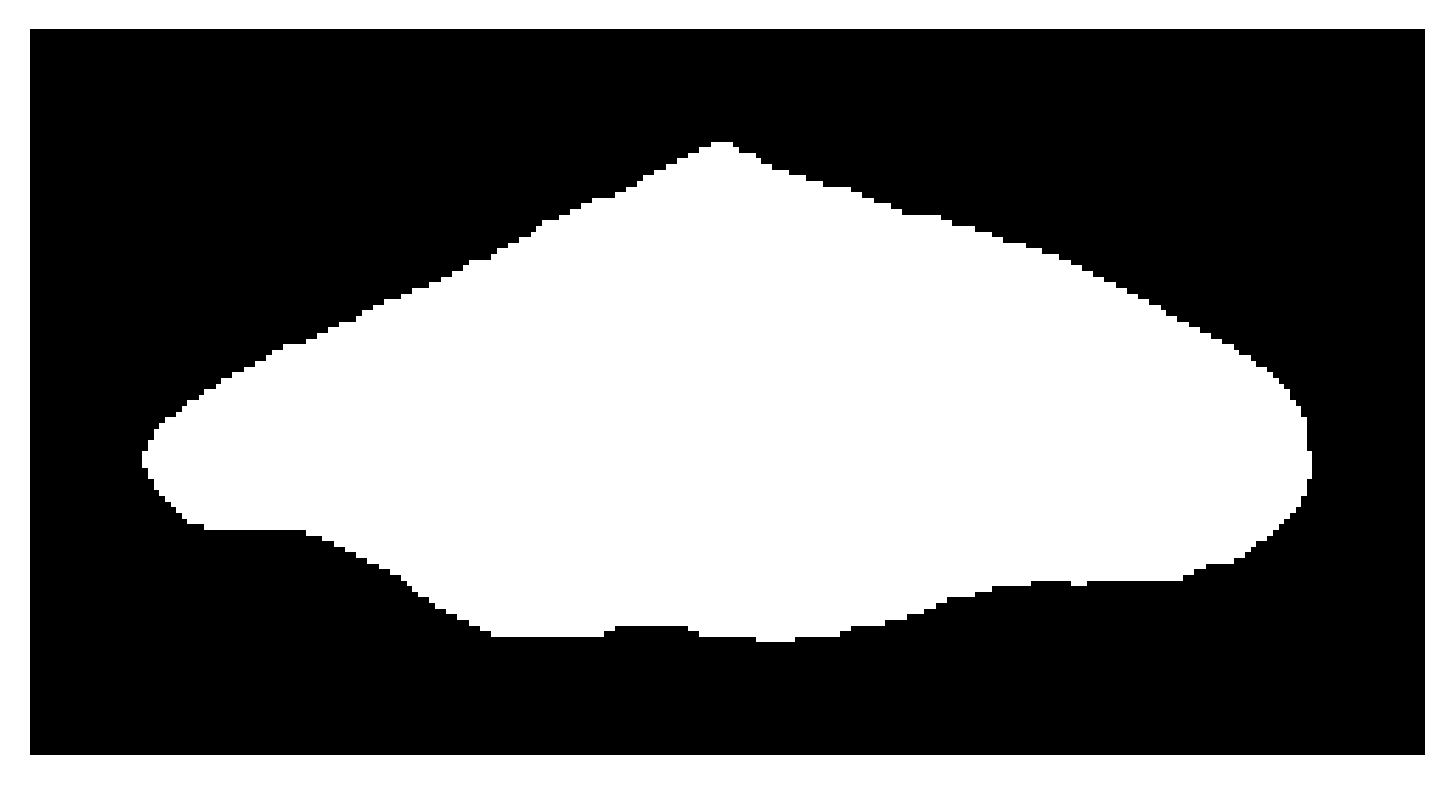}
         \caption{Example image from Krumbein's chart \cite{krumbein1951a}.}
         \label{fig:r3_s3}
     \end{subfigure}
     \begin{subfigure}[b]{0.26\textwidth}
         \includegraphics[width=\columnwidth]{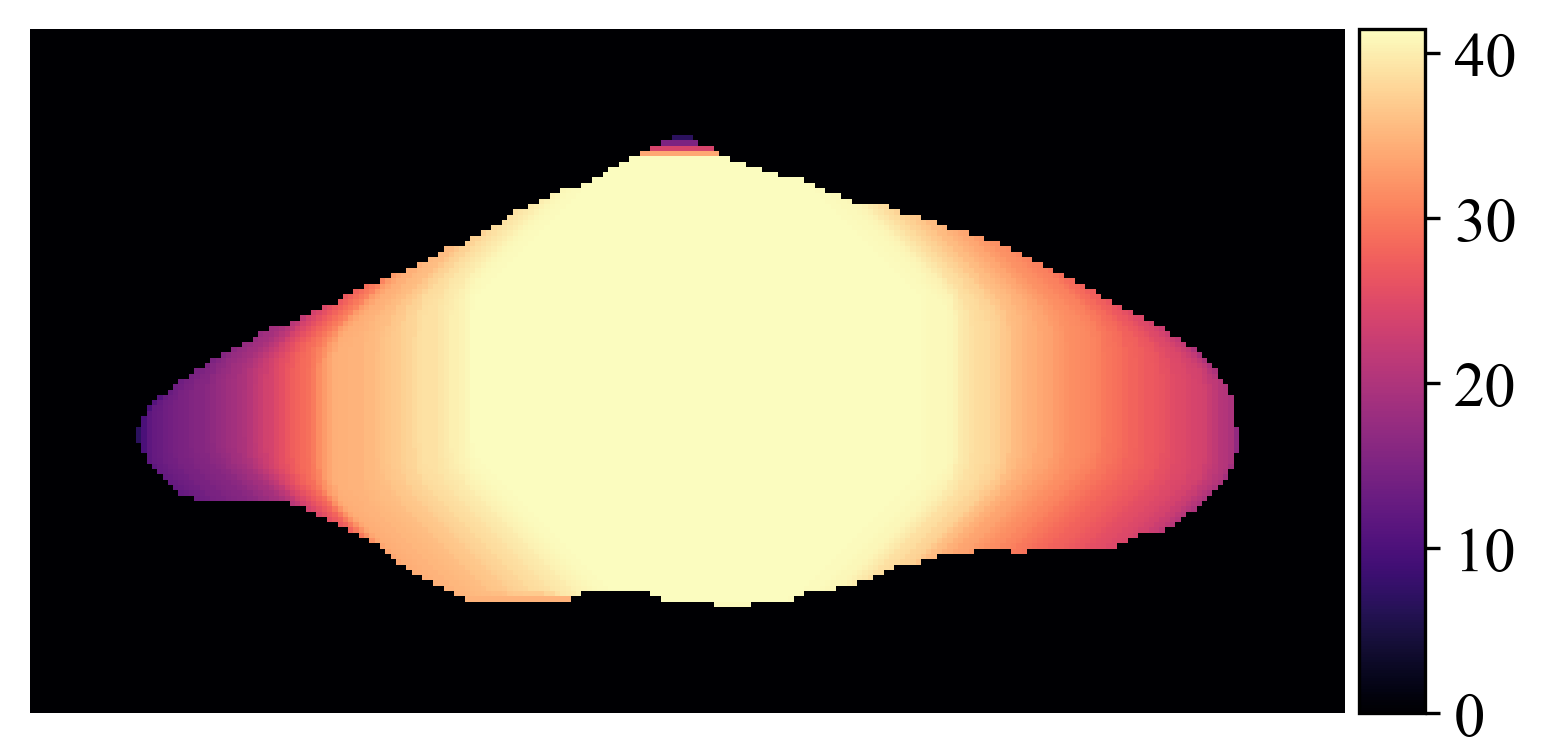}
         \caption{Result of the local thickness calculation.}
         % \vspace{0.32cm}
         \label{fig:r3_s3_thickness}
     \end{subfigure}
     \begin{subfigure}[b]{0.228\textwidth}
         \includegraphics[width=\columnwidth]{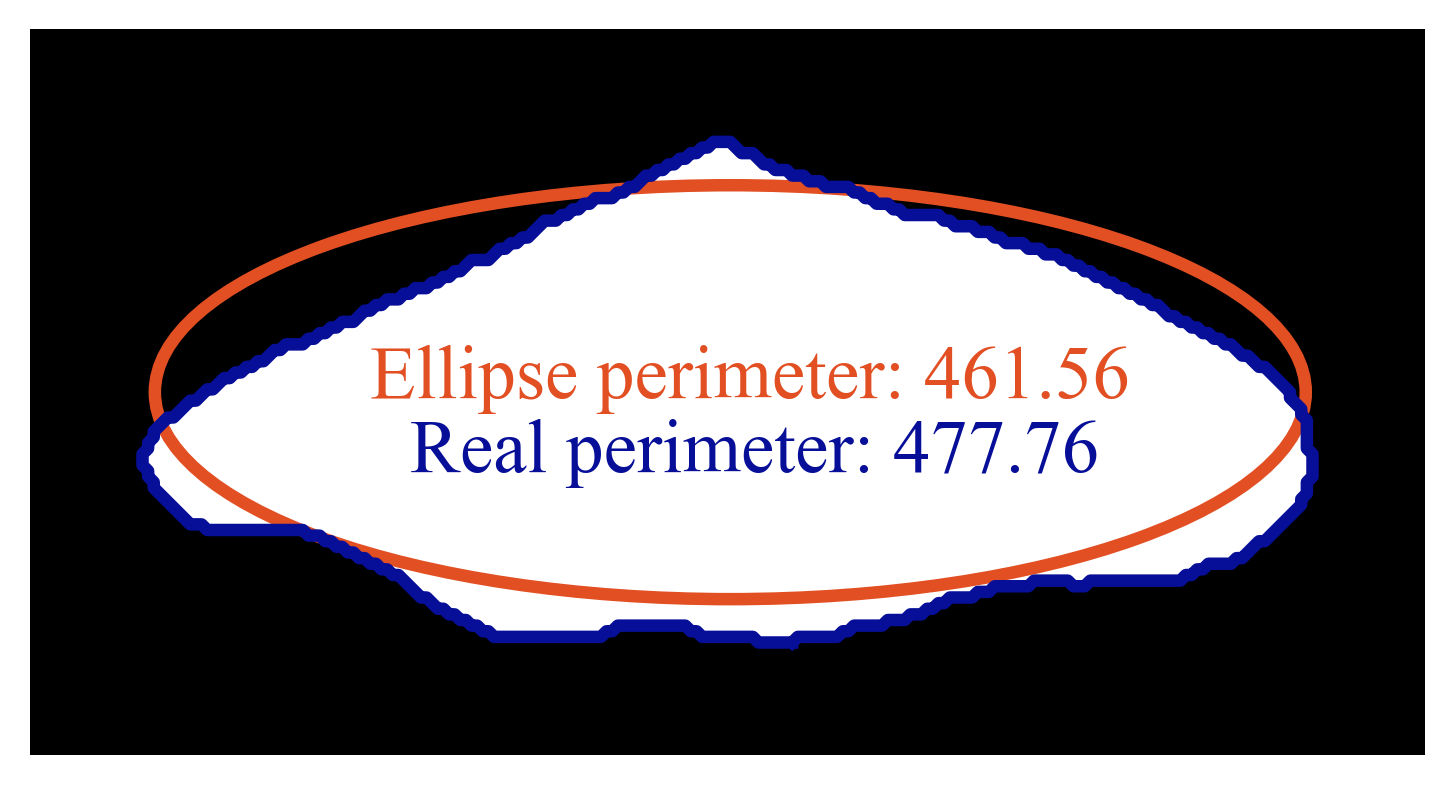}
         \caption{Proposed ellipse-based perimeter and actual object perimeter.}
         \label{fig:r3_s3_perimeter}
     \end{subfigure}
     \begin{subfigure}[b]{0.26\textwidth}
         \includegraphics[width=\columnwidth]{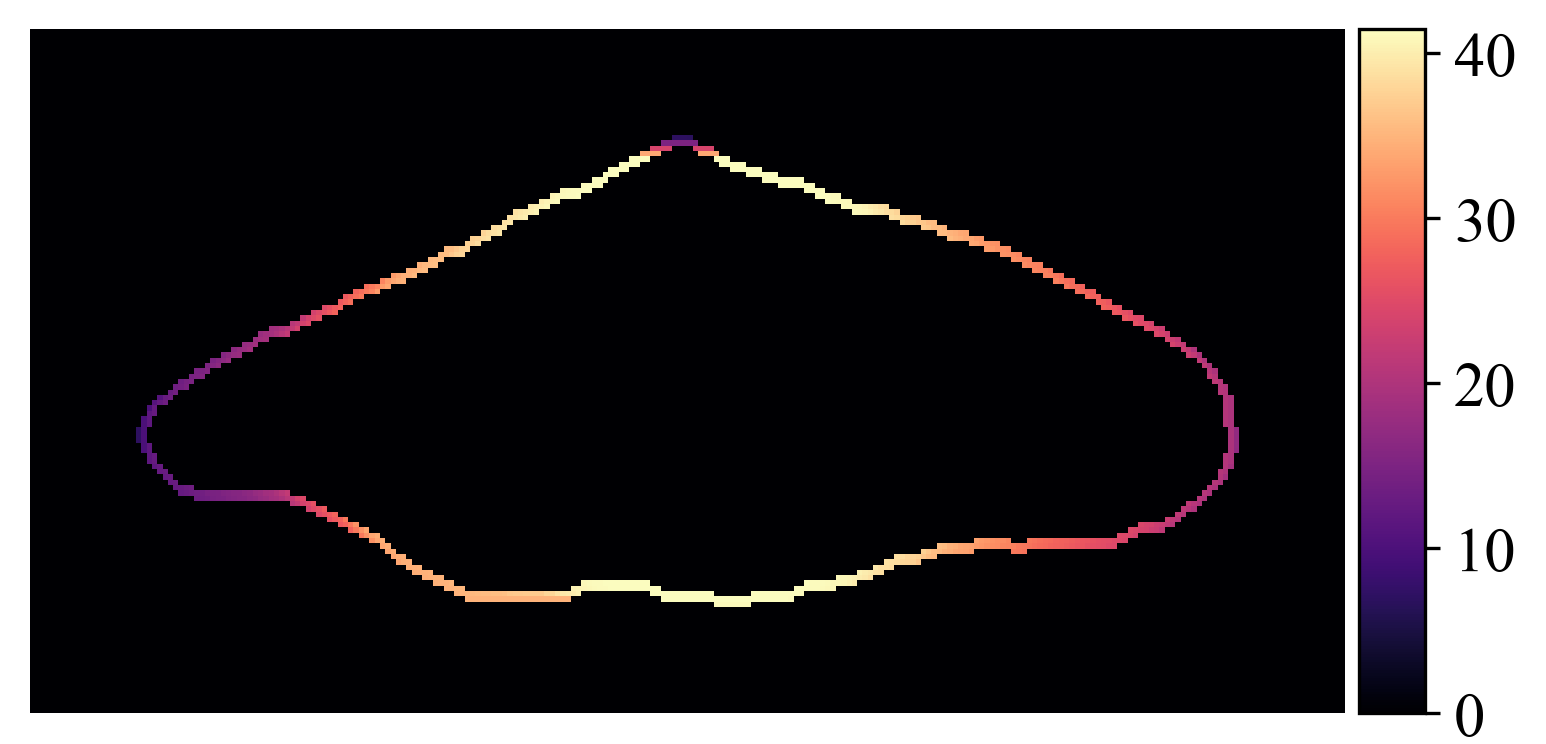}
         \caption{Contour curvature approximation using local thickness values.}
         \label{fig:r3_s3_thickness_edge}
     \end{subfigure}
\caption{Visualization of the components of proposed sphericity and roundness calculation in 2D. Local thickness is calculated on a mask of an object and used to approximate its perimeter and contour curvature.}
\label{fig:method}
\end{figure*}

Local thickness in image processing is defined as the radius of the largest circle (in 2D) or sphere (in 3D) that fits inside an object at any given point~\cite{imagej_localthickness} (\cref{fig:r3_s3_thickness}). Historically, it was calculated by dilating a circle/sphere with incremental radii. However,  more recent work by \citet{dahl2023a_localthickness} proposed an alternative computation approach, reducing the time complexity from $\mathcal{O}(x^7)$ to $\mathcal{O}(x^4)$ for 3D images ($x$ representing the width of a cubic object). 

In our method, we observe that the comprehensive representation of local thickness contains all the necessary information to extract or approximate roundness and sphericity both for 2D and 3D objects. The final time complexity of the proposed methods closely follows that of the local thickness calculation, as the additional steps are mostly limited to a few mathematical operations that can be applied to all objects in a volume simultaneously.

\subsection{Sphericity}
For 3D sphericity, rather than directly calculating the surface area $S$, we propose modeling the object as a spheroid (ellipsoid with two equal semi-diameters). The surface area can then be calculated as: 
\begin{align}
\text{if }& c_s < a_s: \notag\\
    &S = 2\pi a_s^2 \left(1+ \frac{1-e^2}{e}\arctanh{e}\right), \enspace e=1-\frac{c_s^2}{a_s^2} \\
\text{if }& c_s \geq a_s: \notag\\
    &S = 2\pi a_s^2 \left(1+ \frac{c_s}{a_se}\arcsin{e}\right), \enspace e=1-\frac{a_s^2}{c_s^2} 
\end{align}
where $a_s$ is the length of the two equal semi-axes, which we approximate using the mean local thickness. The remaining semi-axis length $c_s$ is calculated from the object volume $V$ (obtained by summing the number of voxels in the object mask):
\begin{equation}
    c_s = \frac{3V}{4 \pi a^2}
\end{equation}

A corresponding modification can also be made for 2D sphericity $\mathcal{S}_{2D,P}$, by using an ellipse instead of a spheroid. In this case, the perimeter is calculated using Ramanujan's first approximation~\cite{villarino2005ramanujansperimeterellipse}:
\begin{equation}
    P_o = \pi \left(3(a_e+b_e) -\sqrt{(3a_e+b_e)(a_e+3b_e)}\right)
\end{equation}
where $a_e$ is the first ellipse axis, again approximated through the mean local thickness, and the second axis $b_e$, as well as $P_o$ from \cref{eq:s2d_p}, are calculated from object area $A$:
\begin{align}
& b_e = \frac{A}{\pi a_e}\\
& P_c = 2\sqrt{\pi A}
\end{align}
An example of the ellipse-based approximation and subsequent perimeter measure is shown on \cref{fig:r3_s3_perimeter}.

\subsection{Roundness}

For roundness, we first recognize that the radius of the maximum inscribed circle is equivalent to the maximum local thickness in an object. Additionally, the sphere-fitting nature of local thickness implies that its values along the object's contour approximate the curvature of that contour. While this approximation is not perfect -- particularly in smooth, elongated regions where the measured curvature is constrained by the value of local thickness -- it offers distinct advantages for roundness calculation:
\begin{itemize}
\item Since local thickness approximates curvature most closely in the corners, it aligns with the corner-oriented definition of roundness,
\item Since the approximated curvature cannot exceed the radius of the maximum inscribed sphere, the calculated roundness cannot exceed 1.
\end{itemize}

The pixel-based contour is extracted by counting object pixels that have a background neighbor in a 4-connectivity neighborhood. This process is fast and can be applied to the entire image at once. The resulting masks are then used to collect the mean of the appropriate local thickness values (\cref{fig:r3_s3_thickness_edge}).

Proposed method can also be seamlessly adapted for 3D objects. Instead of using a contour, mean local thickness values are collected from the object surface using a corresponding 3D neighbor-counting approach.

Similar to the works of \citet{drevin2002a} and \citet{roussillon2009a}, the resulting measure does not follow the $0-1$ range of original roundness but has the necessary properties to correlate closely with it. 
\section{Experiments}
\label{sec:experiments}

We evaluate the proposed methods in both 2D and 3D by benchmarking them against baseline charts and the most recognized existing methods, measuring correlation, and comparing execution speeds.

\subsection{Data}

\begin{figure}[!t]
\centering
     \begin{subfigure}[b]{0.4\columnwidth}
         \includegraphics[width=\columnwidth]{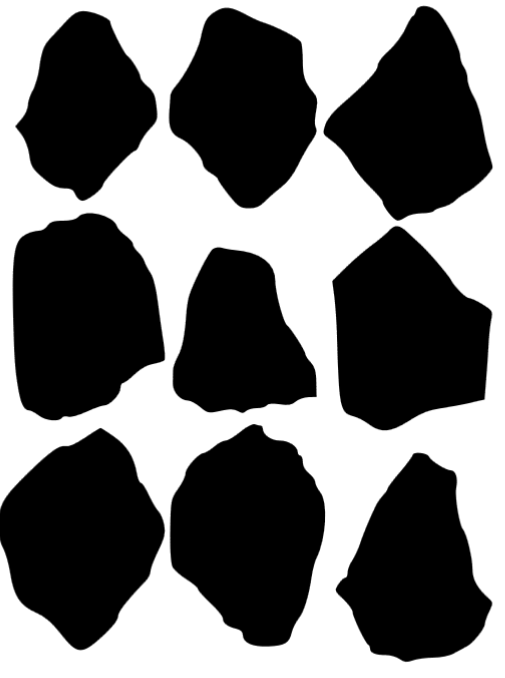}
         \caption{Group of 9 objects with $\mathcal{R}_{\mathrm{2D}}=0.2$.}
         \label{fig:r_collection_2}
     \end{subfigure}
     \hspace{0.4cm}
     \begin{subfigure}[b]{0.4\columnwidth}
         \includegraphics[width=\columnwidth]{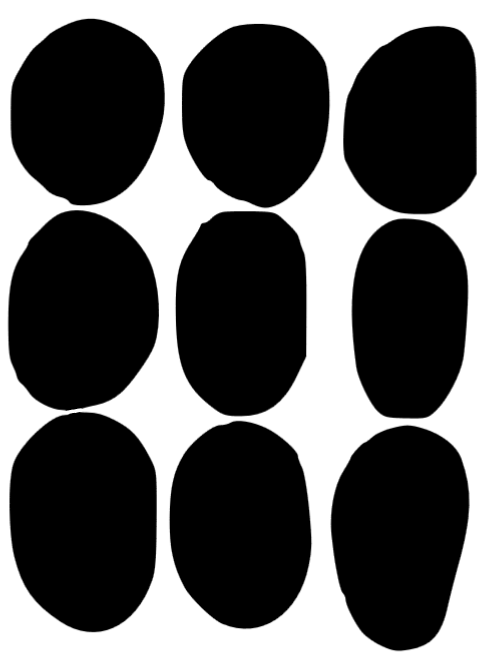}
         \caption{Group of 9 objects with $\mathcal{R}_{\mathrm{2D}}=0.8$.}
         \label{fig:r_collection_8}
     \end{subfigure}
\caption{Sample groups from Krumbein's roundness chart~\cite{krumbein1941a}. The whole chart consists of nine such groups with object roundness values ranging from $\mathcal{R}_{\mathrm{2D}}=0.1$ to $\mathcal{R}_{\mathrm{2D}}=0.9$.}
\label{fig:krumbein_roundess_chart}
\end{figure}

For the initial 2D evaluation, we use two Krumbein charts. The first, introduced in \cref{fig:kumbrein_plot}, contains 20 objects with suggested sphericity and roundness values. The second chart~\cite{krumbein1941a} includes 81 objects, divided into 9 groups with varying roundness values in the range $[0.1,0.2,\dots,0.9]$ (reference samples shown in \cref{fig:krumbein_roundess_chart}). Since roundness has traditionally been a more challenging measure, the second image was often used to compare correlation from an average of 9 measurements --- this is also what is done here.

\begin{figure}[!t]
\centering
     \begin{subfigure}[b]{0.49\columnwidth}
         \includegraphics[width=\columnwidth]{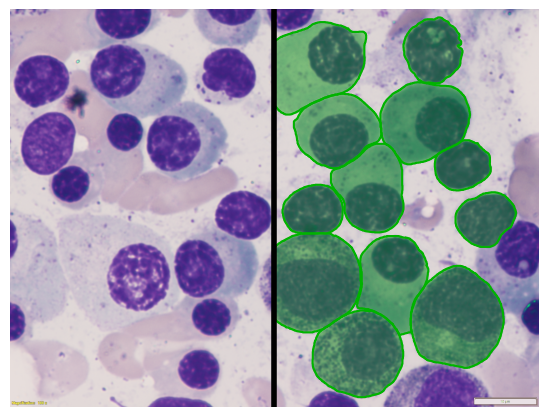}
     \end{subfigure}
     \begin{subfigure}[b]{0.49\columnwidth}
         \includegraphics[width=\columnwidth]{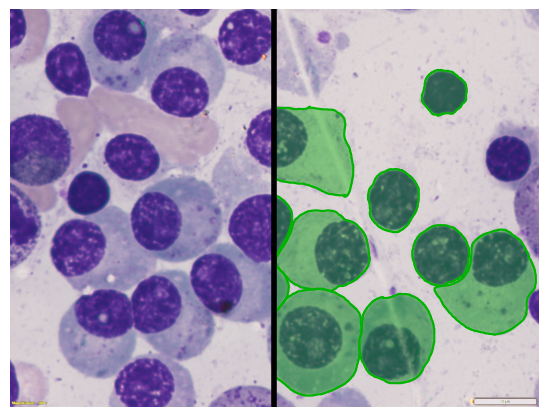}
     \end{subfigure}
          \begin{subfigure}[b]{0.49\columnwidth}
         \includegraphics[width=\columnwidth]{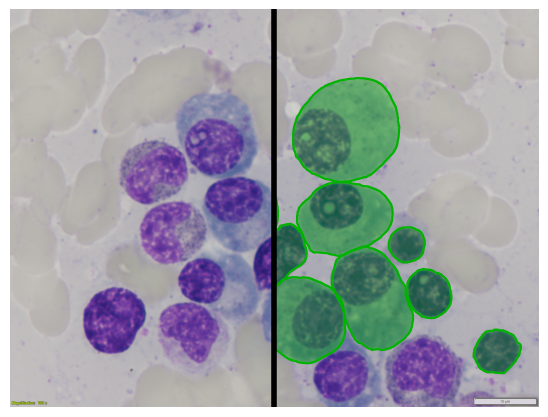}
     \end{subfigure}
          \begin{subfigure}[b]{0.49\columnwidth}
         \includegraphics[width=\columnwidth]{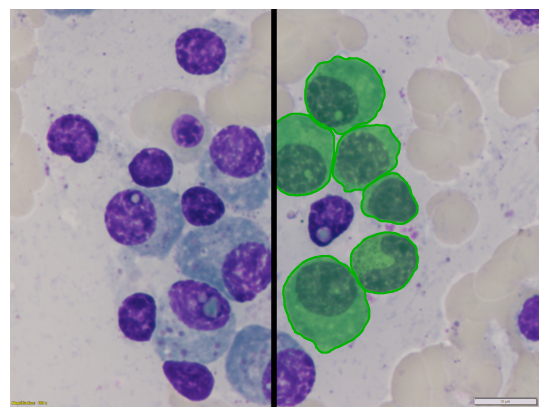}
     \end{subfigure}
\caption{Brightfield microscopy images of cells used in 2D experiments, sourced from~\cite{ma2024a}. Right half of each image visualizes the cell segmentation masks. Edge cells are removed for preservation of realistic shapes.}
\label{fig:micro_2D_data}
\end{figure}

We also evaluate our method on a set of four brightfield cell microscopy images from \emph{The Multi-modality Cell Segmentation Challenge}~\cite{ma2024a} (\cref{fig:micro_2D_data}). Given that cell shape analysis is a common research target~\cite{yin2014a,boquet-pujadas2021a,deng2020a,dulaimi2018a,lesty1989a,kalinin2018a}, these images provide a realistic test case for the proposed methods.

\begin{figure}[!t]
\centering
\includegraphics[width=0.99\columnwidth]{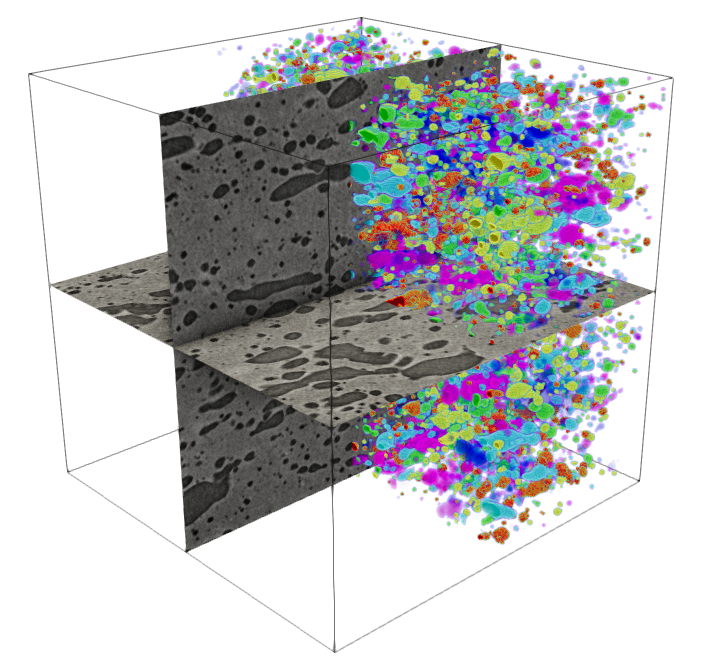}
\caption{Slices from the CT scan of mozzarella microstructure used in the study, together with a fragment of segmented fats (assigned random colors for visualization).}
\label{fig:example_mozz}
\end{figure}

For a test in 3D we use a synchrotron X-ray $\mu$CT scan of mozzarella cheese~\cite{mozzavid}. The microstructure of mozzarella consists of an anisotropic protein matrix interlaced with fats of varying size and complexity~\cite{feng2021a}. By segmenting and labeling each fat component, we obtain a 400-cube volume containing \num{15503} objects, with individual volumes ranging from 27 to approximately \num{130000} pixels (\cref{fig:example_mozz}).

\subsection{Benchmark methods}

We benchmark our method against the most recognized algorithms for exact sphericity and roundness calculations. For both 2D and 3D sphericity, we use the method based on marching cubes/squares~\cite{lin2005a, lorensen1987a}. In particular, we use functions \verb|measure.marching_cubes| and \verb|measure.mesh_surface_area| from the \emph{scikit-image package}~\cite{van2014scikit}, with default parameter settings.

For 2D roundness, we evaluate the method proposed by \citet{zheng2015a}. For that purpose, we use a Python implementation shared in the \emph{wadell\_rs} package~\cite{wadell_rs_git}. Minor parameter fine-tuning is applied for optimal performance, tailored to each dataset.

For 3D roundness, we target the recent 3D extension of the 2D method \cite{zheng2021a} using our own implementation, as the original code is not available. While minor parameter adjustments are occasionally necessary, we decided to keep them consistent with the original paper. Given potential implementation discrepancies, for this method it is more applicable to compare the execution speeds rather than the roundness values.

For conciseness, the proposed local thickness-based sphericity and roundness methods are abbreviated as $\mathcal{S}_{\mathrm{LT}}$ and $\mathcal{R}_{\mathrm{LT}}$, respectively. The marching cubes/squares method is named $\mathcal{S}_{\mathrm{MC}}$, and the 2D/3D roundness methods are named $\mathcal{R}_{\mathrm{Zheng}}$.

We also compare our 2D roundness score against values reported for two approximation methods by \citet{roussillon2009a}, that is their own method (abbreviated here as $\mathcal{R}_{\mathrm{Rouss}}$), and a method by \citet{drevin2002a} ($\mathcal{R}_{\mathrm{Drevin}}$). However since these methods are purely 2D, and their implementation is not available, the investigation is not pursued any further.

\subsection{Computation speed tests}

The computation speed is evaluated in two steps. First, we examine how execution time scales with the size of the objects, both for 2D and 3D data. Importantly, since both measures in the proposed algorithm can share the same local thickness data as input, a simultaneous calculation of both measures is also evaluated.

Next, for 3D data, we evaluate the impact of processing varying numbers of objects simultaneously, rather than focusing on individual object size. The second evaluation process is outlined as follows:
\begin{enumerate}
    \item Sample 10 values for the number of objects (ranging from 1 to \num{2000}) using a logarithmic distribution.
    \item For each sample size, select randomly without replacement 40 times from the \num{15503} available objects to minimize potential biases caused by object volume variation.
    \item For each random selection, compute the necessary measures and record the mean execution time for each method at each sample size.
\end{enumerate}

\section{Results}
\label{sec:results}

\subsection{Performance on 2D charts}

\begin{figure}[!t]
\centering
\includegraphics[width=0.99\columnwidth]{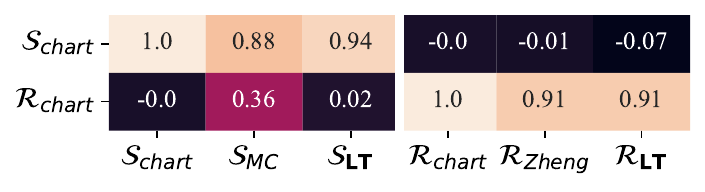}
\caption{Correlation between roundness and sphericity values from Krumbein's chart (\cref{fig:kumbrein_plot}) and evaluated methods. }
\label{fig:krumbein_corr}
\end{figure}

All tested methods exhibited the expected behavior on the first of the Krumbein charts (\cref{fig:krumbein_corr}). The proposed sphericity method ($\mathcal{S}_{\mathrm{LT}}$) achieved a correlation of 0.94, outperforming the baseline ($\mathcal{S}_{\mathrm{MC}}$) which reached a correlation of 0.88. For roundness, both tested methods performed similarly well demonstrating a correlation of 0.91.

Importantly, apart from achieving a high correlation with the respective chart values, the two measure types were also expected to remain uncorrelated with each other. This expectation was met for most methods, except for the baseline sphericity which exhibited a weak correlation of 0.36 with chart roundness.

Results from the second, roundness-only chart (\cref{tab:Krumbein_round_corr}) indicate even better performance than in~\cref{fig:krumbein_corr}. However, reported numbers are only true for correlating mean roundness values within object groups, and not separate objects. Comparing the individual correlations, our method ($\mathcal{R}_{\mathrm{LT}}$) outperforms both the exact baseline ($\mathcal{R}_{\mathrm{Zheng}}$) and one of the approximate methods ($\mathcal{R}_{\mathrm{Drevin}}$), being beaten only by the method proposed by \citet{roussillon2009a} ($\mathcal{R}_{\mathrm{Rouss}}$).

\begin{table}[t!]
\caption{Correlation of various 2D roundness methods with the Krumbein roundness chart (\cref{fig:krumbein_roundess_chart}). Reported values represent the correlation of mean roundness values from objects in each chart group. The underlined values were calculated by us, while the remaining ones were sourced from~\cite{roussillon2009a}.}
\centering
\begin{tabular}{@{}l@{\hskip 5pt}|ccccc@{}}
\toprule
Method & $\mathcal{R}_{\mathrm{Zheng}}$ & $\mathbf{\mathcal{R}_{\mathrm{LT}}}$ & $\mathcal{R}_{\mathrm{Rouss}}$ & $\mathcal{R}_{\mathrm{Drevin}}$ \\ \midrule
Correlation & \underline{0.964} & \textbf{\underline{0.986}} & 0.992 & 0.967 \\ \bottomrule
\end{tabular}
\label{tab:Krumbein_round_corr}
\end{table}

\begin{figure}[!t]
\centering
\includegraphics[width=0.99\columnwidth]{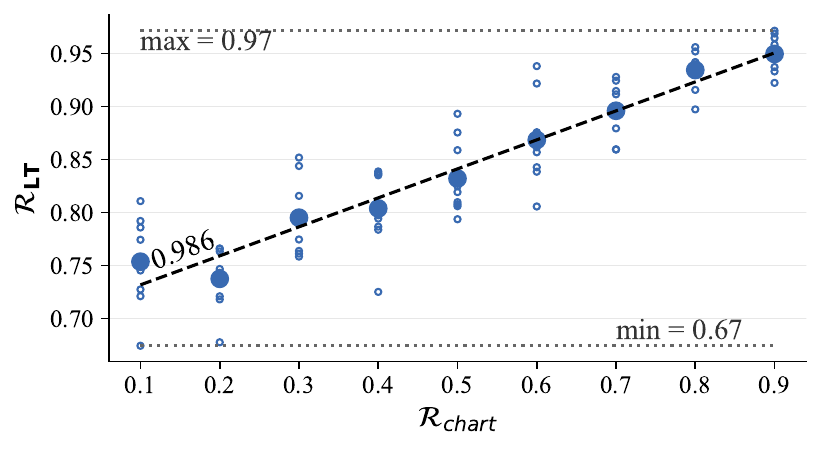}
\caption{Output of the proposed roundness method in comparison to the Krumbein roundness chart values. Mean value is shown with a large filled circle, individual samples are shown as open small circles. Regression line and correlation are shown in black.}
\label{fig:krumbein_round_res}
\end{figure}

A closer analysis of the method output (\cref{fig:krumbein_round_res}) indicates its consistent performance, especially for objects of higher roundness. Slightly higher variability is observed for low-roundness objects, likely caused by incorporating information from the entire contour rather than just the corners. As expected, the maximum reported value does not exceed 1, and the minimum never reaches the 0 value. If necessary, the reported minimum could be used to map the roundness scores to the standard 0--1 range.

\subsection{2D and 3D data}

\begin{figure*}[!t]
\centering
    \begin{subfigure}[b]{0.49\textwidth}
    \includegraphics[width=\columnwidth]{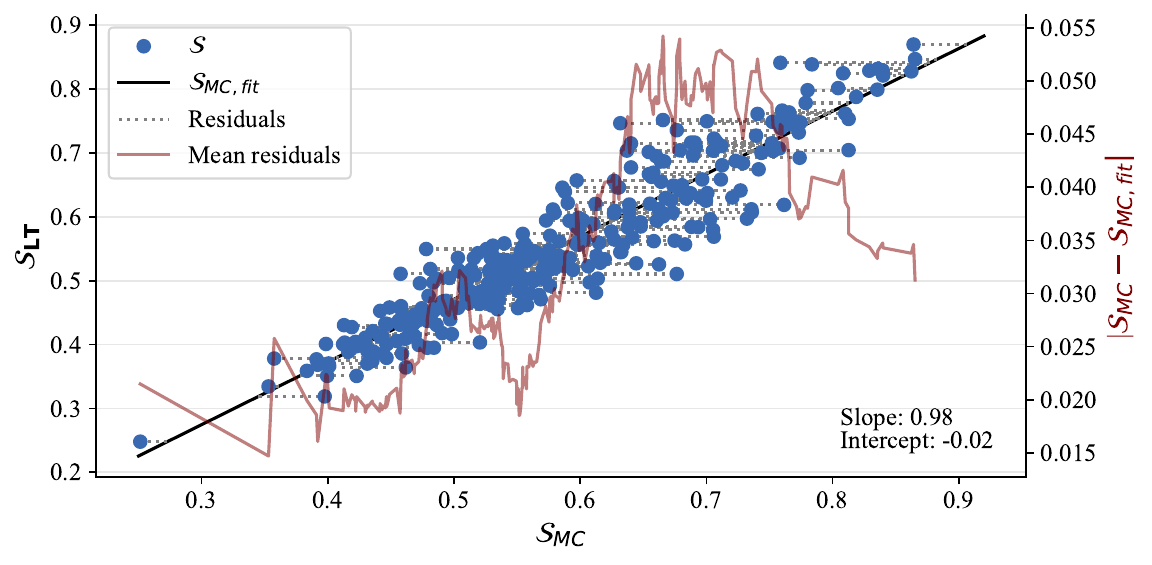}
        \caption{Sphericity measures and their correlation on a subset of biggest objects from the 3D mozzarella scan (\cref{fig:example_mozz}).}
        \label{fig:mozz_spher_acc}
    \end{subfigure}
    \begin{subfigure}[b]{0.49\textwidth}
    \includegraphics[width=\columnwidth]{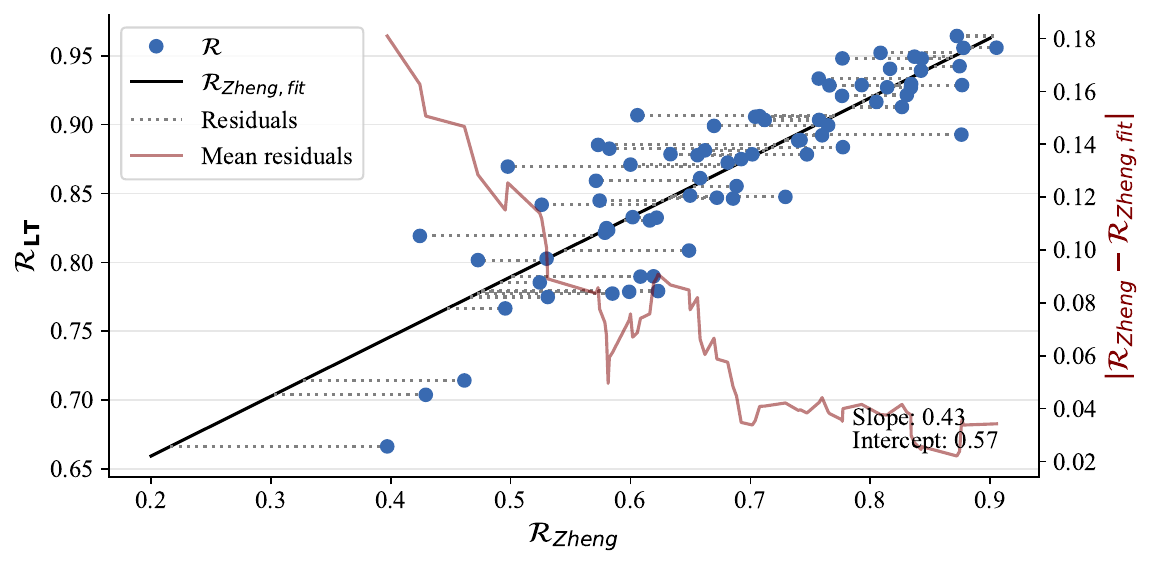}
        \caption{Roundness measures and their correlation on nuclei from the 2D microscopy data (\cref{fig:micro_2D_data}).}
        \label{fig:zheng_round_acc}
    \end{subfigure}
\caption{Correlation between the baseline and proposed methods on the 2D and 3D data. In both cases, linear regression is fit to the data, and residuals are calculated for each data point. }
\label{fig:acc_res_3D}
\end{figure*}

Sphericity methods are further investigated based on results from the 3D mozzarella data. For that purpose, the smallest objects are filtered out ($V<10^3$), as their measurement will likely be distorted. A regression fit calculated for remaining objects ($\mathcal{S}_{\mathrm{MC,fit}}$) shows a very strong correlation between the two measures, with small residuals across the sphericity range (\cref{fig:mozz_spher_acc}). Furthermore, the values of slope and intercept highlight a clear trend similarity between the two measures. The running mean suggests that the difference tends to be bigger for objects with high sphericity, but this effect is still relatively small, rarely exceeding 0.1.

For roundness, the analysis focuses on the 2D data, due to the uncertainty caused by a lack of an established 3D definition and implementation. The resulting fit ($\mathcal{R}_{\mathrm{Zheng,fit}}$) shown on~\cref{fig:zheng_round_acc} suggests a clear correlation between the two measures. However, the discrepancy is higher than for sphericity, with some residuals reaching values of almost 0.2. This is especially true for objects of lower roundness, where the methods tend to disagree the most.

\begin{figure}[!t]
\centering
     \begin{subfigure}[b]{0.4\columnwidth}
     \includegraphics[width=\columnwidth]{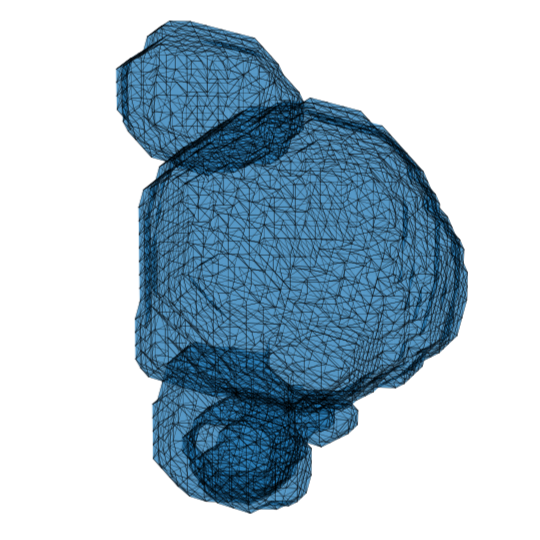}
        \caption{$V$ = 6229, $\mathcal{S}_{\mathrm{MC}}$ = 0.63,\\$\mathcal{S}_{\mathrm{MC,fit}}$ = 0.78}
        \label{fig:mozz_3d_vis_1}
     \end{subfigure}
     \begin{subfigure}[b]{0.4\columnwidth}
     \includegraphics[width=\columnwidth]{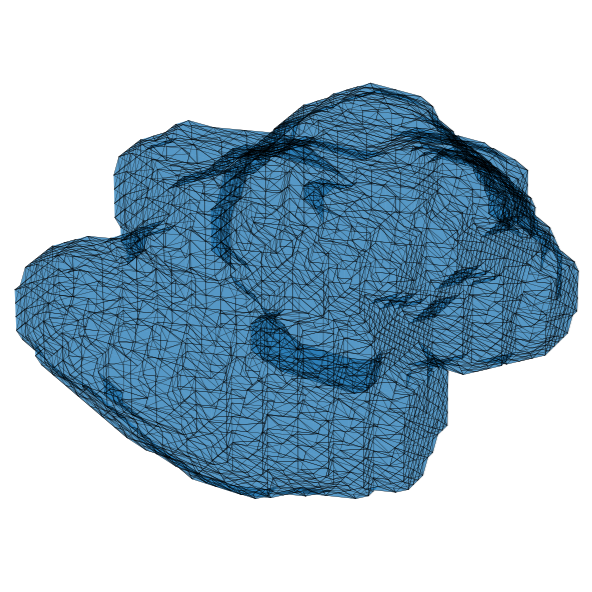}
        \caption{$V$ = 6399, $\mathcal{S}_{\mathrm{MC}}$ = 0.68,\\$\mathcal{S}_{\mathrm{MC,fit}}$ = 0.54}
        \label{fig:mozz_3d_vis_2}
     \end{subfigure}
     \begin{subfigure}[b]{0.4\columnwidth}
     \includegraphics[width=\columnwidth]{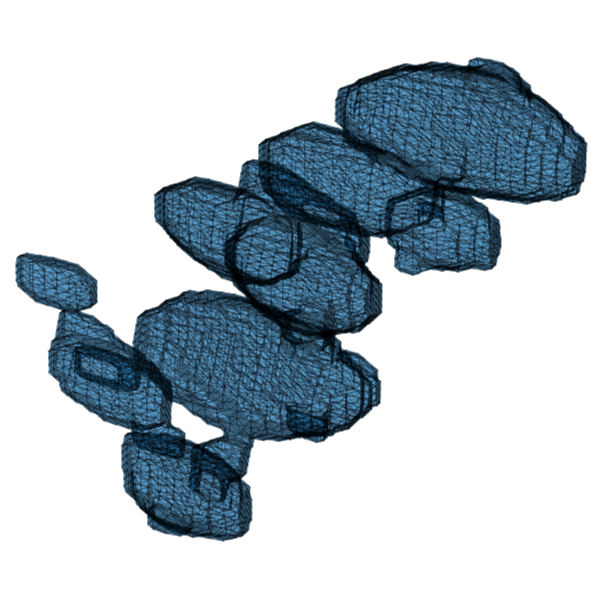}
        \caption{$V$ = \num{16566}, $\mathcal{S}_{\mathrm{MC}}$ = 0.42,\\$\mathcal{S}_{\mathrm{MC,fit}}$ = 0.42}
        \label{fig:mozz_3d_vis_3}
     \end{subfigure}
     \begin{subfigure}[b]{0.4\columnwidth}
     \includegraphics[width=\columnwidth]{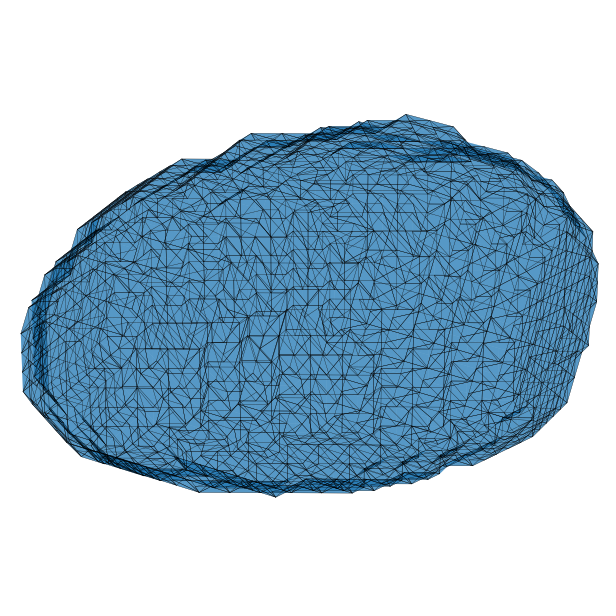}
        \caption{$V$ = 4006, $\mathcal{S}_{\mathrm{MC}}$ = 0.84,\\$\mathcal{S}_{\mathrm{MC,fit}}$ = 0.83}
        \label{fig:mozz_3d_vis_4}
     \end{subfigure}
\caption{Visualization of example 3D objects used for evaluating sphericity methods. Top row consists of two objects where the compared methods showed the biggest discrepancy. Bottom row shows objects with the lowest discrepancy.}
\label{fig:mozz_3d_vis}
\end{figure}

\begin{figure}[!t]
\centering
     \begin{subfigure}[b]{0.45\columnwidth}
     \includegraphics[width=0.96\columnwidth]{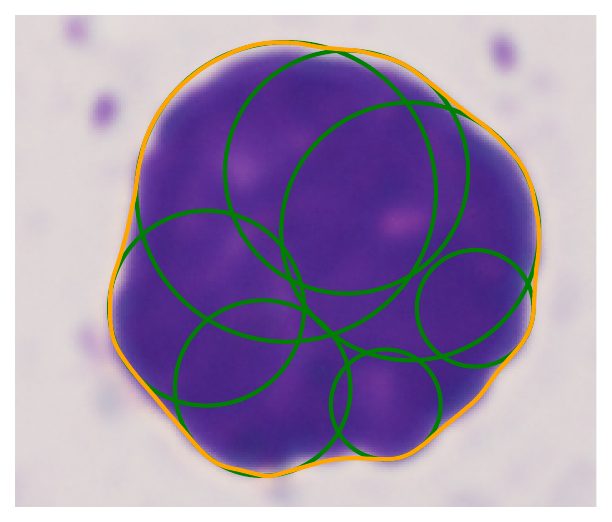}
        \caption{$\mathcal{R}_{\mathrm{Zheng}}$ = 0.498,\\$\mathcal{R}_{\mathrm{Zheng,fit}}$ = 0.685}
        \label{fig:zheng_rvis_1}
     \end{subfigure}
     \begin{subfigure}[b]{0.45\columnwidth}
     \includegraphics[width=\columnwidth]{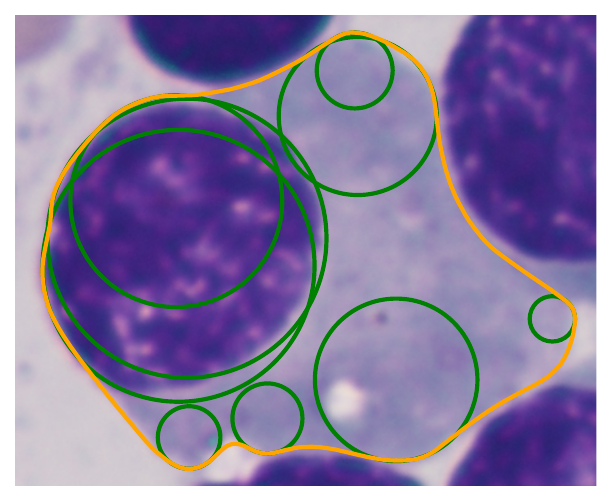}
        \caption{$\mathcal{R}_{\mathrm{Zheng}}$ = 0.397,\\$\mathcal{R}_{\mathrm{Zheng,fit}}$ = 0.216}
        \label{fig:zheng_rvis_2}
     \end{subfigure}
     \begin{subfigure}[b]{0.45\columnwidth}
     \includegraphics[width=\columnwidth]{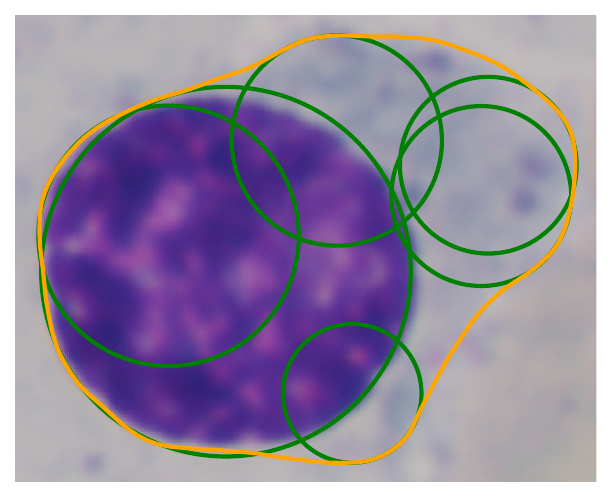}
        \caption{$\mathcal{R}_{\mathrm{Zheng}}$ = 0.601,\\$\mathcal{R}_{\mathrm{Zheng,fit}}$ = 0.600}
        \label{fig:zheng_rvis_3}
     \end{subfigure}
     \begin{subfigure}[b]{0.45\columnwidth}
     \includegraphics[width=0.98\columnwidth]{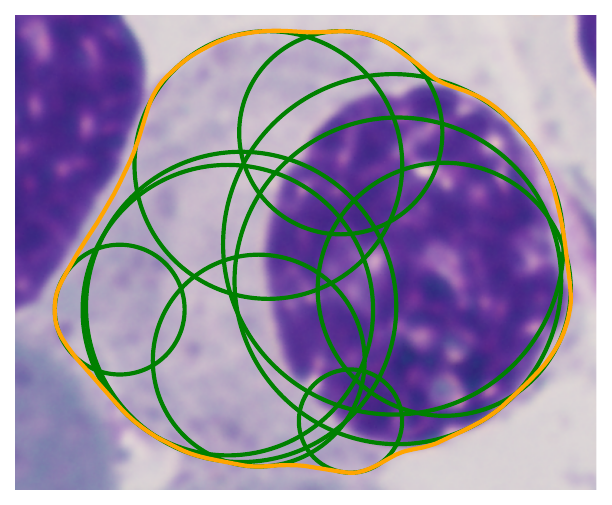}
        \caption{$\mathcal{R}_{\mathrm{Zheng}}$ = 0.580,\\$\mathcal{R}_{\mathrm{Zheng,fit}}$ = 0.582}
        
        \label{fig:zheng_rvis_4}
     \end{subfigure}
\caption{Visualization of example cell images used for evaluating roundness methods. Top row consists of two cells where the compared methods showed the biggest discrepancy. Bottom row shows cells with the lowest discrepancy. Green circles represent corners detected by the $\mathcal{R}_{\mathrm{Zheng}}$ method. Orange line draws a smoothed boundary of the nucleus.
}
\label{fig:zheng_rvis}
\end{figure}

The examples with the largest and smallest residuals between the baseline methods and the fit line are shown in \cref{fig:mozz_3d_vis} and \cref{fig:zheng_rvis}. For sphericity, these examples align with the mean residual trend in~\cref{fig:mozz_spher_acc}. The highest discrepancies occur for moderately complex objects with medium-high sphericity (\cref{fig:mozz_3d_vis_1,fig:mozz_3d_vis_2}). For an object with low sphericity (\cref{fig:mozz_3d_vis_3}), the difference is minimal. Similarly, for a non-spherical object (\cref{fig:mozz_3d_vis_4}), the two measures are closely aligned, despite its complex shape.

Examples from the roundness evaluation provide a potential explanation for the higher discrepancy reported in \cref{fig:zheng_round_acc}. In the first high-residual case (\cref{fig:zheng_rvis_1}), the baseline method fits a large number of corner circles to very minor bends, resulting in a low roundness score. In contrast, the second example has two protruding and sharp corners, but the baseline method fails to capture their significance, as it can only assign one value per corner. The resulting difference in $\mathcal{R}_{\mathrm{Zheng}}$ value between these two images is only 0.1, despite the nucleus in~\cref{fig:zheng_rvis_1} being significantly rounder. The two positive examples (\cref{fig:zheng_rvis_3,fig:zheng_rvis_4}) are harder to interpret, but suggest that the methods align best when the baseline approximates the overall contour curvature and not just discrete corner values.

\subsection{Execution time evaluation}

\begin{figure*}[!t]
\centering
     \begin{subfigure}[b]{0.33\textwidth}
         \includegraphics[width=\columnwidth]{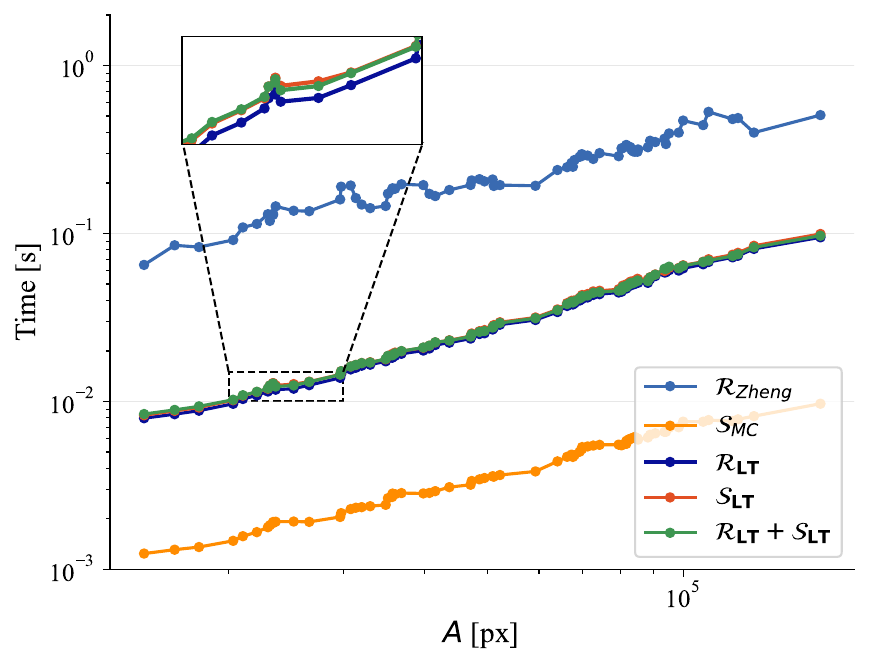}
        \caption{Execution time on separate 2D objects of varying area from \cref{fig:micro_2D_data}.}
        \label{fig:zheng_speed}
     \end{subfigure}
     \begin{subfigure}[b]{0.33\textwidth}
        \includegraphics[width=0.99\columnwidth]{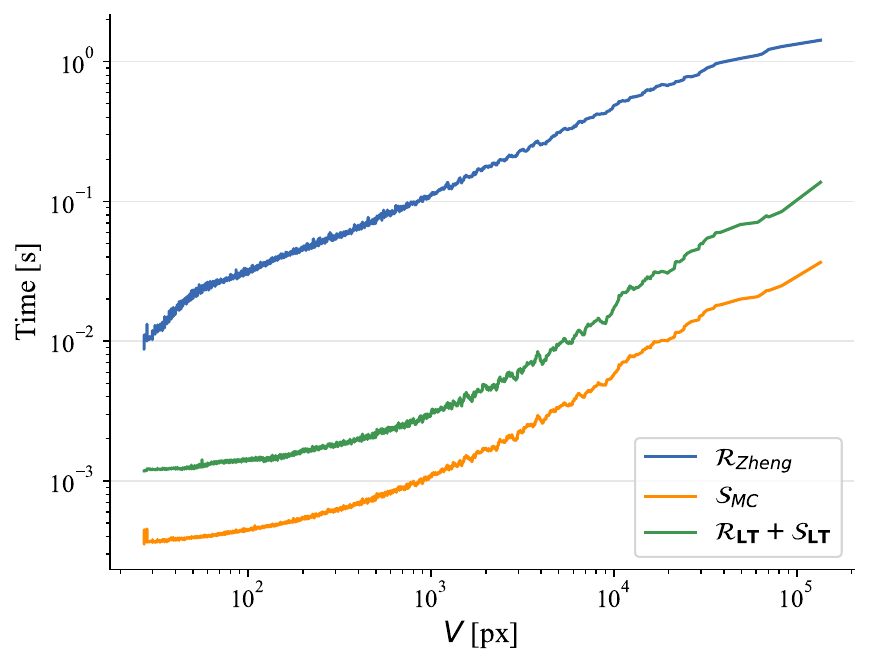}
        \caption{Execution time on separate 3D fats from the mozzarella scan.}
        \label{fig:mozz_speed_one_obj}
     \end{subfigure}
     \begin{subfigure}[b]{0.33\textwidth}
        \includegraphics[width=0.99\columnwidth]{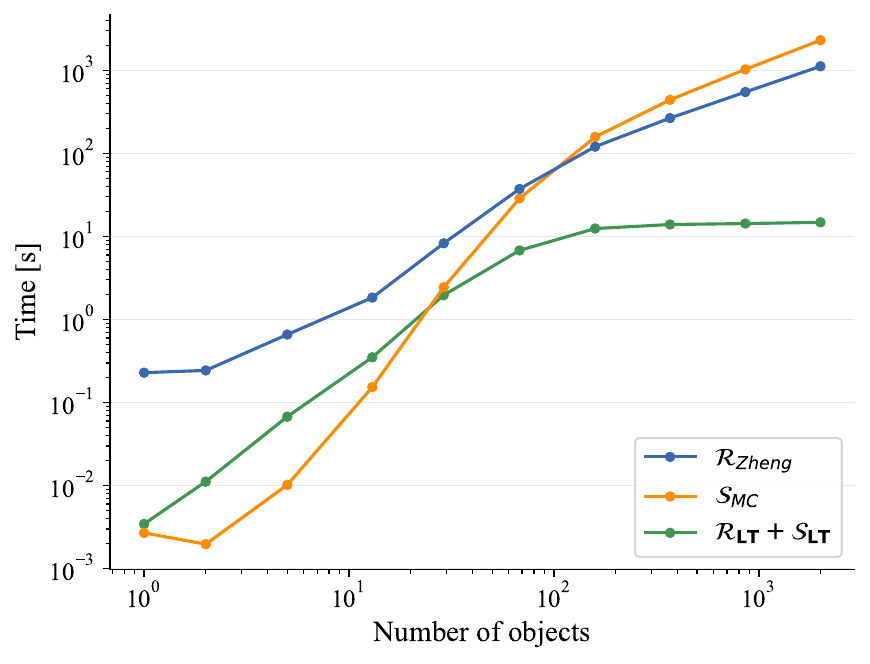}
        \caption{Execution time for evaluating multiple 3D fats from the mozzarella scan at once.}
        \label{fig:mozz_speed_multi_obj}
     \end{subfigure}
\caption{Execution time of investigated methods on 2D and 3D data. The first 2D test reports both the execution time for separate methods and for calculating the two measures together using the same local thickness output. The following figures report only the combined time, as it closely approximates the cost for each separate calculation.}
\label{fig:speed_res}
\end{figure*}

The execution time measured on 2D data (\cref{fig:zheng_speed}) reveals that the proposed methods are faster than the baseline roundness, but also noticeably slower than $\mathcal{S}_{\mathrm{MC}}$, when evaluating single objects of varying area.

This trend continues for 3D data (\cref{fig:mozz_speed_one_obj} ) but is inverted when calculating not single objects but whole groups of objects (\cref{fig:mozz_speed_multi_obj}). The proposed methods scale very well with increasing object count, significantly outperforming both sphericity and roundness baselines. For instance, calculating baseline sphericity for 2,000 objects took 38 minutes, while the proposed method needed only 14 seconds.

The zoomed-in section of~\cref{fig:zheng_speed} shows that computing sphericity, roundness, or both measures at once has nearly identical time complexity, confirming that the execution speed of the proposed methods is primarily influenced by the underlying local thickness algorithm.

\section{Discussion and Limitations}
\label{sec:discussion}

\subsection{Performance}

The performance of the proposed methods demonstrates their effectiveness compared to traditional, exact approaches. In the chart-based tests, they achieve similar, or better results than tested baselines and other approximate methods. Furthermore, their fast execution, especially when processing multiple objects at once opens the potential for fast image-based statistical analysis, that was previously unattainable.

For sphericity, the reported values closely match the baseline, even for irregular and complex shapes. Some discrepancies can be observed for medium-sphericity objects, but they are too minor to compromise the method's applicability. In the case of roundness, the observed discrepancies are more pronounced, but a big part of that can be attributed to the disputable output of the baseline $\mathcal{R}_{\mathrm{Zheng}}$ method.

\subsection{Applicability of roundness approximations}

While methods for exact calculation of roundness have only recently become available, many approximations have been known for a very long time. However, they have not gained widespread adoption, likely due to their perceived inferiority to exact measures.

Contrary to this trend, our observations suggest that in most image analysis tasks an exact roundness measurement is often unnecessary, or even unfeasible. Already in 2D, accurately fitting circles to corners is challenging due to resolution limitations and the influence of roughness, requiring a careful choice of method parameters. In 3D, the process is further complicated by the unresolved influence of ridges.

We propose that assessing the curvature of the entire contour or surface is a more practical approach. It effectively captures corner curvature while avoiding the ambiguity caused by parameter choice and enabling a seamless application of roundness on 3D data. We also expect that with contour/surface methods, the influence of roughness can be easily controlled through various filtering approaches.

\subsection{Similarity to the method proposed by Drevin and Vincent}

Some parallels can be drawn between our method and the theory presented by~\citet{drevin2002a} who developed the $\mathcal{R}_{\mathrm{Drevin}}$ method analyzed in~\cref{tab:Krumbein_round_corr}. They explored the effect of morphological opening on the object mask, noting its potential to express roundness, sphericity, and roughness. By performing iterative opening with increasing disk radius, they achieve a representation that superficially corresponds to the one shown in local thickness. They further propose that sphericity can be approximated by calculating the area removed from the object by performing an opening with a disk kernel of radius $0.42R$.

Despite the similarities, their method has many practical limitations. The ratio of $0.42R$ is based on a small set of examples and was later questioned by~\citet{roussillon2009a} who suggested that higher ratios may yield more consistent results. Additionally, the execution time of binary opening is very sensitive to the size of the structural element, with a time complexity of $\mathcal{O}(x^3 s^3)$, where $s$ is the width of the structural element. This means that for ratios larger than $0.05R$, such morphological opening will be slower than a local thickness calculation.
\section{Conclusion}
\label{sec:conclusion}

Calculating roundness and sphericity based on the local thickness algorithm provides a fast and accurate alternative to the exact methods. The proposed implementations perform well both on 2D and 3D data, occasionally outperforming the baselines in correlating to the roundness and sphericity on test charts. With a single local thickness output, both measures can be computed simultaneously for thousands of objects within a significantly shorter time than with existing approaches. Implementation of the methods is available for use in Python, as a pip-installable package under \url{https://github.com/PaPieta/fast_rs}.

\vspace{0.4cm}
\noindent \textbf{Acknowledgments}: This work was supported by Innovation Fund Denmark, project 0223-00041B (ExCheQuER).
\clearpage
{
    \small
    \bibliographystyle{ieeenat_fullname}
    \bibliography{main}
}

\end{document}